\documentclass[conference]{IEEEtran}
\IEEEoverridecommandlockouts

\def\IEEEtitletopspaceextra{0.25in}

\usepackage{graphicx}
\usepackage{amsmath}
\usepackage{tabularx}
\usepackage{amsfonts}
\usepackage{amssymb}
\usepackage{gensymb}
\usepackage{booktabs}
\RequirePackage[colorlinks=true, allcolors=blue]{hyperref}
\usepackage{orcidlink}  

\usepackage[
    backend=biber,
    style=ieee,
    natbib=true,
    doi=false,
    url=false,
    isbn=false,
]{biblatex}

\AtEveryBibitem{%
   \clearfield{Title}%
   \clearfield{eprint}%
   \clearfield{note}%
}

\usepackage{etoolbox}

\makeatletter
\patchcmd{\@maketitle}{\Huge}{\LARGE}{}{}
\makeatother

\AtBeginEnvironment{tabular}{\scriptsize}
\renewcommand{\citet}[1]{\textcite{#1}}
\renewcommand{\citep}[1]{\parencite{#1}}

\title{BiomechAgent: AI-Assisted Biomechanical Analysis Through Code-Generating Agents}

\author{
    \IEEEauthorblockN{1\textsuperscript{st        }R. James Cotton \orcidlink{0000-0001-5714-1400} }
    \IEEEauthorblockA{
            \textit{Shirley Ryan AbilityLab}\\       \textit{Northwestern University}                  \\
      rcotton@sralab.org
          } \and     \IEEEauthorblockN{2\textsuperscript{nd        }Thomas Leonard}
    \IEEEauthorblockA{
            \textit{Stritch School of Medicine}\\       \textit{Loyola University Chicago}                  \\
      tleonard1@luc.edu
          }}

\begin{document}

\maketitle


\begin{abstract}
Markerless motion capture is making quantitative movement analysis increasingly accessible, yet analyzing the resulting data remains a barrier for clinicians without programming expertise. We present BiomechAgent, a code-generating AI agent that enables biomechanical analysis through natural language and allows users to querying databases, generating visualizations, and even interpret data without requiring users to write code. To evaluate BiomechAgent's capabilities, we developed a systematic benchmark spanning data retrieval, visualization, activity classification, temporal segmentation, and clinical reasoning. BiomechAgent achieved robust accuracy on data retrieval and visualization tasks and demonstrated emerging clinical reasoning capabilities. We used our dataset to systematically evaluate several of our design decisions. Biomechanically-informed, domain-specific instructions significantly improved performance over generic prompts, and integrating validated specialized tools for gait event detection substantially boosted accuracy on challenging spatiotemporal analysis where the base agent struggled. We also tested BiomechAgent using a local open-weight model instead of a frontier cloud based LLM and found that perform was substantially diminished in most domains other than database retrieval. In short, BiomechAgent makes the data from accessible motion capture and much more useful and accessible to end users.\\
\end{abstract}


\section{Introduction}

Progress in markerless motion capture is rapidly making quantitative motion analysis much cheaper and easier to obtain \citep{kanko_concurrent_2021, uhlrich_opencap_2023, cotton_differentiable_2025}. It can now even be integrated into clinical settings and detect meaningful differences in gait \citep{cimorelli_validation_2024, peiffer_portable_2025}. This is essential for overcoming a critical barrier in clinical movement rehabilitation: \textit{we typically cannot measure what we treat}. While walking speed has been proposed as the sixth vital sign and is commonly measured with a stop watch \citep{fritz_white_2009}, the details of how people move carries substantial additional information; information that can now be quantitatively captured.

However, scalable access to motion analysis creates a new challenge: analyzing all this data. For commonly performed activities, such as gait analysis, high-quality, activity-specific algorithms are available.
For example, we developed the GaitTransformer \citep{cotton_transforming_2022, cimorelli_validation_2024}, which can accurately detect gait events from video alone. However, there are many other cases where novel questions require novel analysis. Even navigating a database to identify specific trials and writing code to visualize kinematics during these trials is a barrier to wider use, and formal biomechanical analysis is outside the training scope of most rehabilitation clinicians.

Our recent work BiomechGPT \citep{yang_biomechgpt_2025} partially addresses this challenge by developing a multimodal large language model (MLLM) that is fluent in both biomechanics and language. It is trained to answer clinically relevant tasks in several domains. For example, it shows high performance at classifying activities or detecting impaired gait. However, it can only answer the classes of questions for which it is trained. Furthermore, like many language models, it can make errors (hallucinations), and the interpretability challenge means that it cannot explain the rationale for its answers.

As an alternative approach, LLMs are rapidly improving their performance on coding challenges \citep{austin_program_2021, jimenez_swe-bench_2024, jiang_survey_2025}, including potentially biomechanical analysis and interpretation. Furthermore, LLMs can be equipped with tools that enable them to interact with external data sources or users, giving rise to agentic systems \citep{yao_react_2023, schick_toolformer_2023, wang_agentic_2025}. Vision language models (VLMs) can also view and help interpret figures they produce \citep{alayrac_flamingo_2022, liu_visual_2023, masry_chartqa_2022, liu_picture_2025}. BiomechAgent is such an agent, with tools for retrieving biomechanics data from a database, writing code to analyze it, generating plots, visualizing these plots, and packaging it into a familiar and intuitive chat interface.


Essential to this ability is the fact that all of our data is processed in PosePipe \citep{cotton_posepipe_2022}, which leverages DataJoint \citep{yatsenko_datajoint_2015} to track all of our data collection and processed results in a database. In this work, we focused on analyzing biomechanics obtained with our multiview markerless motion capture system \citep{cotton_optimizing_2023, cotton_improved_2023, cotton_differentiable_2025}, but this is naturally extensible to biomechanics obtained from our monocular Portable Biomechanics Laboratory \citep{peiffer_portable_2025}.

We see BiomechAgent and BiomechGPT as synergistic in the same way as the cognitive science theory of System 1 and System 2 thinking \citep{kahneman_thinking_2011}. Specifically, System 1 thinking is an automatic, nearly effortless form of perception (e.g., BiomechGPT directly outputs an answer upon observing biomechanical sequences). In contrast, System 2 is effortful and involves logical reasoning (e.g., BiomechAgent writes code and explicitly manipulates data). Furthermore, BiomechAgent can be provided with access to other biomechanical analysis tools, such as BiomechGPT.

Our goal in this work is to systematically benchmark BiomechAgent and measure its performance across multiple domains to understand its performance limits better. This can guide future improvements and inform end users of the tasks for which it can be trusted. However, a challenge for benchmarking agentic systems is that while the potential uses of BiomechAgent are open-ended, quantitative and interpretable benchmarking requires defining a discrete set of tasks. To address this, we started with open-ended user testing to identify some example use cases. From these, we established several domains of tasks, including targeted data retrieval, data visualization, activity classification and segmentation, and even clinical reasoning. For many of these, we had additional ground-truth data that we used to develop our benchmarking dataset, including clinical reasoning and activity classification, tested in a two-alternative forced-choice (2AFC) paradigm. In some cases where it is harder to define an exact answer, we used an LLM-as-a-Judge framework to evaluate the performance of BiomechAgent \citep{zheng_judging_2023}.

We also tested the importance of several of our design decisions, including custom instructions for our agent that provide context for biomechanical analysis and the use of LLMs that can be run locally (Gemma models) versus frontier, cloud-based models (Gemini 2.5 Flash Lite). In addition, we tested whether providing the agent with specialized analysis tools, such as gait event detection using our GaitTransformer \citep{cotton_transforming_2022, cimorelli_validation_2024}, improved performance on tasks on which it otherwise struggled.

Overall, we found that BiomechAgent is a powerful tool that provides a user-friendly, convenient chat interface for both data retrieval and plotting. It also demonstrated impressive examples of reasoning about clinically meaningful features of movement with only modest prompting. In short, our contributions are:

\begin{itemize}
\item We developed BiomechAgent, an agentic system for accessing and analyzing biomechanical data collected with markerless motion capture to facilitate this analysis by clinicians.
\item We developed an evaluation dataset that was motivated by tasks we wanted BiomechAgent to solve, where we could validate the answer either directly against data unavailable to BiomechAgent or using an LLM-as-a-Judge.
\item We demonstrated that BiomechAgent shows strong performance on many of these tasks, and validated some of our design decisions, including using a cloud-based frontier model (Gemini 2.5 Flash Lite), custom instructions to the agent about biomechanical analysis, and providing additional tools to the agent for detecting gait event timing.
\end{itemize}

\section{Methods}

\subsection{BiomechAgent System}

\subsubsection{Agent Architecture}

BiomechAgent is a code-generating agent architecture built on the smolagents framework \citep{smolagents}. Unlike function-calling approaches, in which language models output structured JSON to invoke predefined functions, smolagents generates and executes Python code directly, which is effective for expressing agentic plans \citep{zheng_judging_2023}. The agent receives a natural-language query, reasons about which computational steps are required, writes executable code, observes the results in a sandboxed environment, and iterates until it reaches a final answer.


The agent accesses specialized capabilities through special Python functions called tools. These tools provide database access for querying participant trials and fetching kinematic data \citep{yatsenko_datajoint_2015}, gait analysis for detecting foot contact events from kinematics using GaitTransformer \citep{cotton_transforming_2022}, and visualization for generating static and interactive plots. Tool documentation embedded in function definitions informs the agent how to use each tool, including the expected inputs and outputs, and the agent then writes code to flexibly use these tools to retrieve and analyze data.

During execution of BiomechAgent, the agent receives a user prompt. It then begins emitting text, including flags indicating that sections are code. The smolagent framework extracts those from the LLM as they are produced, and when a complete block of code is received, it is then executed. The agent then receives that as the new input during a chat sequence. Once the agent determines it can answer the question, it responds with a `final\_answer' tool call which stops the sequence.

\subsubsection{Language Model Backends}

We evaluated BiomechAgent with two language-model backends, representing cloud and local deployment scenarios. Gemini 2.5 Flash Lite \citep{comanici_gemini_2025}, a frontier cloud-based model, and MedGemma 27B \citep{sellergren_medgemma_2025}, an open-weights medical foundation model, both of which can process images and text. Whenever the agent generated an image and used the `save\_plot' tool, the image was presented to the end user and provided to the visual language model (VLM), along with the text, for subsequent chat iterations.


\subsubsection{Custom Instructions}

Domain-specific guidance is embedded in the agent's system prompt to prevent common errors discovered during early testing. These instructions include biomechanical formulas (e.g., the calculation of the asymmetry index) and expected parameter ranges. It also provides guidance on standards, such as using consistent colors to indicate left and right, using real units of time and degrees on the axes, and avoiding common errors (e.g., forgetting to import mathematical libraries or plotting tools). They also encourage BiomechAgent to reason about clinical questions and to formulate hypotheses to explore in the data, as often observed in the traces provided in the supplementary document.

\subsubsection{Data Sources}

Kinematics were obtained from multiview markerless motion capture \citep{cotton_differentiable_2025}. Data were managed via DataJoint \citep{yatsenko_datajoint_2015} and the PosePipe processing framework \citep{cotton_posepipe_2022}. The Northwestern University Institutional Review Board approved all procedures, and participants provided informed consent.

\subsubsection{Chat Interface}

Our end-user chat interface was implemented in streamlit \citep{streamlit_2024}. The user inputs are provided to BiomechAgent and then the resulting code and plots are shown to the user as they are generated until the agent emits a `final\_answer' in response to the input. At that point, the user can either start a new chat or ask a follow-up question to continue the existing chat interface.

\subsection{Dataset Generation}

We developed test cases across seven categories. These were developed by first mining the authors' transcripts using the system in an open-ended manner and then identifying categories of use. We distilled these into multiple topics within each category, where the correct answers could be checked against other available data. These additional data sources were not made available to BiomechAgent. We generated templates for each of these topics to allow a dataset to be generated for subsequent testing.

\subsubsection{Privileged Data Sources}

A critical design principle is the separation between data sources available to the agent and those used solely for ground-truthing during dataset generation. During evaluation, the agent has access only to kinematic reconstructions and activity annotations derived from video. The following privileged sources generate ground truth answers but are \textbf{unavailable to the agent during evaluation}:

Instrumented walkway: Per-step spatiotemporal measurements (velocity, step length, stride time) serving as ground truth for gait parameter estimation. Clinical records: Diagnosis labels (stroke, prosthesis user, control), affected side, and level of amputation. Berg Balance Scale assessments: Standardized fall risk scores obtained within a week of a gait trial for inpatients. Activity segmentation annotation: Manual labels of activity segmentation, such as rising from a chair, turning, or the push phase of wheelchair propulsion.

\subsubsection{Database Tasks}

Database tasks test the agent's ability to query and aggregate information from the motion capture database. Questions span trial counts (``How many trials does participant X have?''), data shapes (``What are the dimensions of the kinematics array?''), activity queries (``List all TUG trials for participant X''), session information, and date spans. Ground truth is computed via direct queries to the database using validated calls. These tasks are deterministic and verify that the agent correctly interfaces with the database.

\subsubsection{Activity Classification Tasks}

Activity classification tasks present two-alternative forced choice (2AFC) questions requiring the agent to identify which of two trials contains a specific activity---such as backward walking versus forward walking or Timed Up and Go versus overground walking---by analyzing kinematics alone. Success requires identifying discriminative kinematic features: reversed ankle plantarflexion-dorsiflexion sequences for backward walking, characteristic turning patterns for TUG, or altered coordination patterns for different gait conditions. The agent was specifically instructed not to use activity annotations, and the tool providing activity labels was blocked via a forbidden tools constraint to enforce this.

\subsubsection{Spatiotemporal Tasks}

Spatiotemporal tasks evaluate estimation of gait parameters from kinematics without access to instrumented walkway measurements. Topics include velocity and spatial parameters (walking speed, step length, stride length), temporal parameters (stride time, stance and swing duration, stance/swing ratios), gait event detection (identifying stance and swing phase timing), event counting (number of left and right foot contacts), and asymmetry metrics (left-right differences in step length and timing).

\subsubsection{Clinical Tasks}

Clinical tasks require inferring clinical status from kinematic patterns. In a 2AFC format, the agent must determine whether one of two participants has a stroke (versus an able-bodied control) or uses a lower-limb prosthesis, identify the affected side, or compare fall risk. Ground truth comes from clinical records and Berg Balance Scale assessments. Success requires recognizing clinically relevant patterns, such as an asymmetric range of motion.

\subsubsection{Segmentation Tasks}

Segmentation tasks test the identification of temporal phases within complex activities. Questions address phase duration (``How long was the turning phase?''), timing (``When did the sit-to-stand transition begin?''), and counting (``How many walking segments occurred?''). Ground truth comes from manual video annotations. These tasks evaluate whether the agent can parse continuous motion into clinically meaningful segments.

\subsubsection{Visualization Tasks}

Visualization tasks assess the agent's ability to generate appropriate plots for different analytical needs: single joint trajectories, bilateral comparisons, multi-joint summaries, pelvis trajectories, and range of motion summaries. Evaluation uses an LLM-as-judge approach with domain-specific rubrics to assess whether the visualization addresses the request and displays appropriate data.

\subsubsection{Perception Tasks}

Perception tasks evaluate the agent's ability to interpret time series plots and extract meaningful observations. Unlike visualization tasks which test plot generation, perception tasks require the agent to first create a visualization and then accurately describe what it observes. Topics include clinical pattern recognition (detecting ROM asymmetry indicative of stroke or prosthesis use), activity pattern recognition (identifying turning, backwards walking, or head nodding from kinematic traces), movement phase identification (recognizing sit-to-stand transitions, walking onset, or L-shaped trajectories), and gait cycle analysis (estimating cadence, assessing stride symmetry). The key evaluation criterion is whether the agent's description is concordant with the expected pattern given ground truth clinical or activity information that the agent cannot access directly. LLM-as-judge scoring emphasizes specific value citations over generic descriptions.

\subsection{Evaluation}

\subsubsection{Agent Configurations}

We tested five agent configurations to understand the contributions of different components:

\bigskip\noindent
\begin{tabular}{p{\dimexpr 0.30\linewidth-2\tabcolsep}p{\dimexpr 0.70\linewidth-2\tabcolsep}}
\toprule
Configuration & Description \\
\hline
\textbf{Baseline} & Gemini 2.5 Flash Lite with custom biomechanics instructions \\
\textbf{+GaitTransformer} & Baseline with access to learned gait event detection tool \\
\textbf{No Instructions} & Default prompts only, removing domain-specific guidance (lesion study) \\
\textbf{No Vision} & Baseline without feeding generated plots back to the model \\
\textbf{MedGemma} & Local 27B model for on-premises deployment \\
\bottomrule
\end{tabular}

\bigskip

The GaitTransformer variant was the only one with access to this tool; it otherwise matched the Baseline variant and was tested only on the spatiotemporal dataset. The ``No Instruction'' variant lacked our custom biomechanically grounded instructions. The MedGemma variant used a local 27B model running on an A100 with 4-bit quantization.  The ``No Vision'' did not provide images to the LLM when the agent generated them and removed the instructions about looking at them.

\subsubsection{Scoring Methods}

\textbf{Deterministic scoring} was applied where ground truth permits objective evaluation. Categorical responses (activity names, laterality, participant IDs) require exact string matches. Numerical responses are specified with tolerances appropriate to measurement uncertainty: $\pm$10\% relative tolerance for most spatiotemporal parameters (velocity, step length, temporal measures), $\pm$5 percentage points for asymmetry metrics, $\pm$1 for footfall counts, and exact matches for database counts and data shapes.

\textbf{LLM-as-judge} \citep{zheng_judging_2023} was used for tasks requiring semantic evaluation, particularly visualization quality and clinical reasoning. Domain-specific rubrics score multiple dimensions on a 0-10 scale. Importantly, the rubrics evaluate reasoning quality alongside correctness: well-reasoned answers that reach incorrect conclusions score higher than lucky guesses without justification, encouraging genuine analytical capability over pattern matching.


\subsubsection{Evaluation Visualization}

We also developed a Streamlit GUI to visualize agent behavior from the evaluation dataset. This shows a chat interface and the responses from BiomechAgent, including the reasoning traces, the code that was executed, and the plots that were generated.

\subsubsection{Additional Metrics}

Beyond accuracy, we tracked token usage and execution time to characterize computational costs across configurations. We also tracked metrics such as the number of coding errors produced.

%

\section{Results}

\subsection{End user testing}

\begin{figure*}[!htbp]
\centering
\includegraphics[width=1\linewidth]{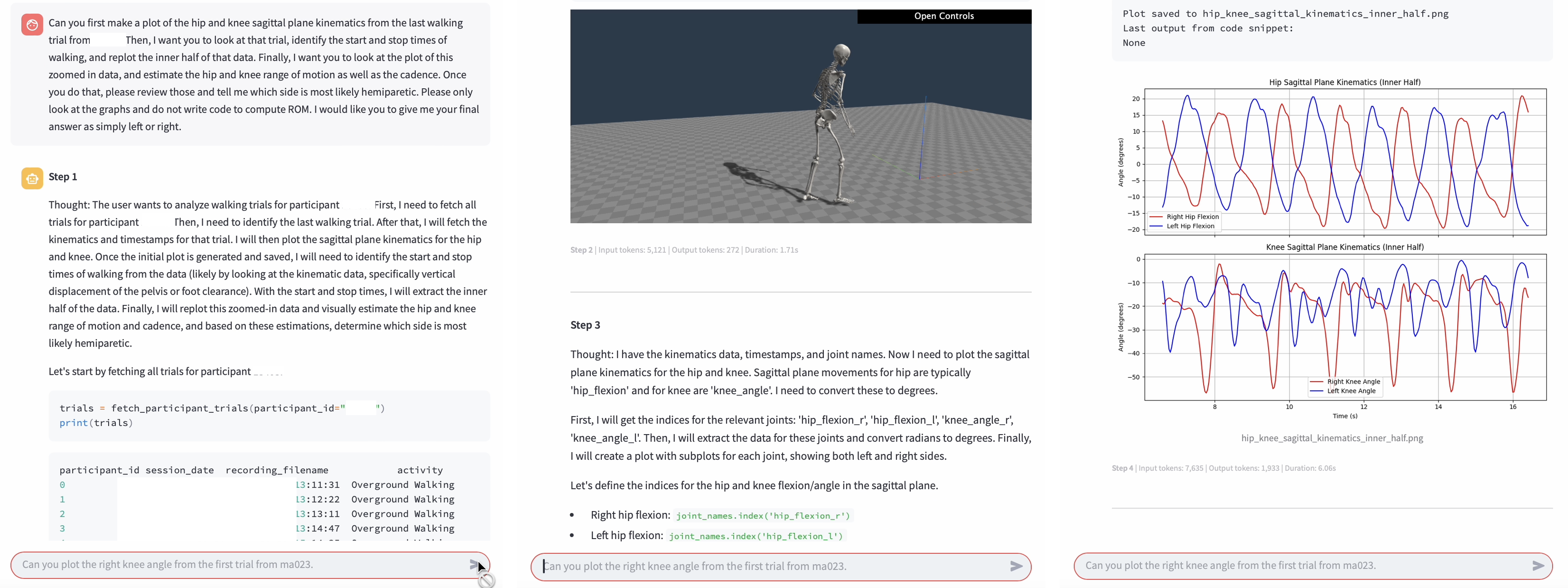}
\caption[]{BiomechAgent chat interface screenshots from a sequence of reasoning about the side with hemiparesis, including plotting and feeding the result into the VLM.}
\label{fig-chat-interface}
\end{figure*}

Initial testing of BiomechAgent was informal, conducted through ad hoc use by the authors, who both found it easy and intuitive to use. It consistently identified trials, filtered them by activity to select the desired ones, and generated plots aligned with our instructions. Figure~\ref{fig-chat-interface} shows an example dialog in which the user asks the system to analyze hip and knee kinematics and determine which side exhibits hemiparesis. Despite never being instructed in this task, it generated a plot, analyzed it using the VLM, and identified the side with hemiparesis. Logs from the initial use of BiomechAgent were then used to develop some of the question classes employed in our systematic benchmarking.

\subsection{Overall Performance between Models}

\begin{figure}[!htbp]
\centering
\includegraphics[width=0.8\linewidth]{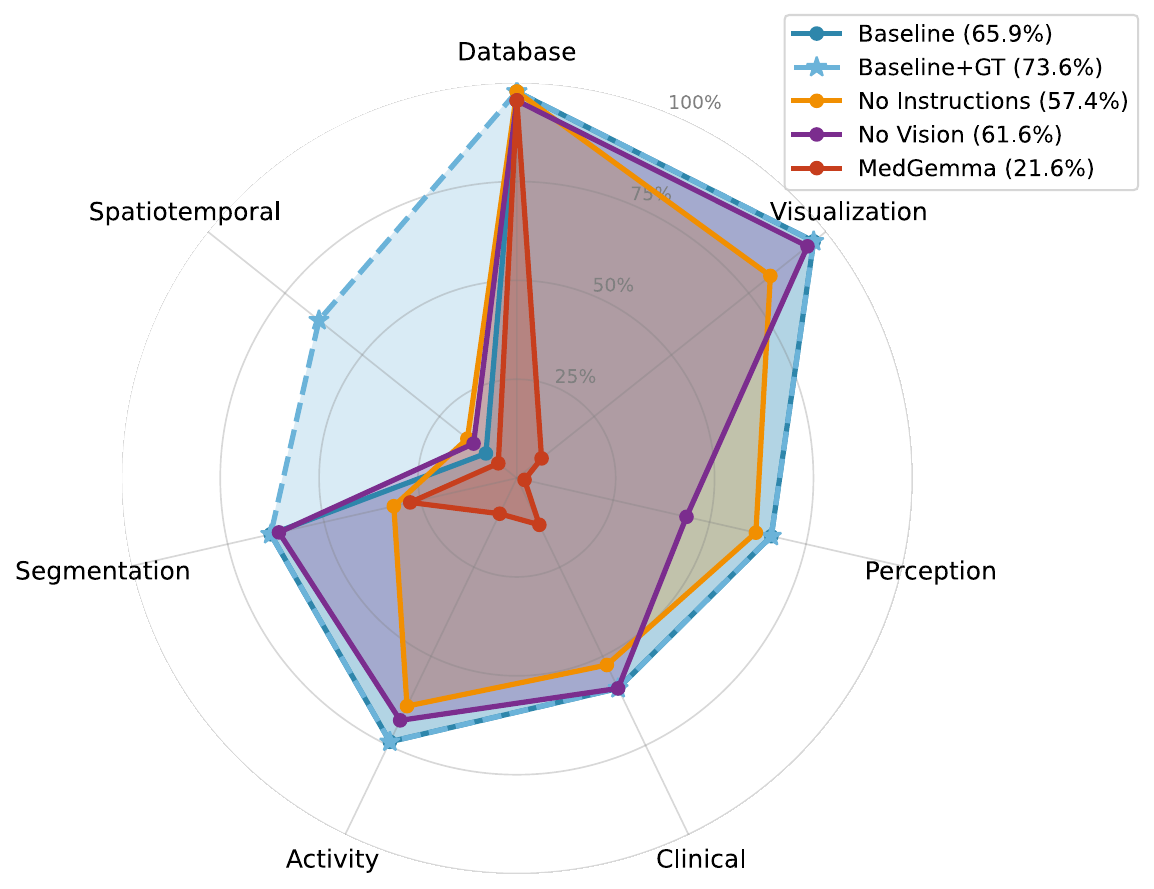}
\caption[]{Performance comparison across task categories for different model configurations. Baseline+GT replaces the spatiotemporal results from the baseline evaluation with those when the GaitTransformer tool was also included.}
\label{fig-radar-comparison}
\end{figure}

Figure~\ref{fig-radar-comparison} shows performance across different domains for each of our agent variations. Our baseline model performed as well as or better than a model without custom prompt instructions (i.e., only the default instructions from smolagents \citep{smolagents}). Using a local model, MedGemma 27B, solved many tasks, but not as well as using Gemini 2.5 Flash Lite. All models performed poorly on spatiotemporal tasks, except when provided access to special gait analysis tools described below. We also discuss the `No Vision' tests in more depth below.


\subsection{Detailed Performance Analysis of Baseline Model}

\begin{table}
\centering
\caption[]{Detailed performance by topic group (Baseline)}

\begin{tabular}{p{\dimexpr 0.333\linewidth-2\tabcolsep}p{\dimexpr 0.333\linewidth-2\tabcolsep}p{\dimexpr 0.333\linewidth-2\tabcolsep}}
\toprule
Category & Topic & Baseline \\
\hline
Database & Database Access & 4/4 (100.0\%) \\
 & Data Shape & 7/8 (87.5\%) \\
 & Activity Context & 8/8 (100.0\%) \\
 & Trial Duration & 4/4 (100.0\%) \\
 & Session Count & 4/4 (100.0\%) \\
 & Joint Index & 4/4 (100.0\%) \\
 & Date Span & 4/4 (100.0\%) \\
 & Activity Diversity & 8/8 (100.0\%) \\
Visualization & (all topics) & 48/50 (96.0\%) \\
Perception & (all topics) & 33/50 (66.0\%) \\
Clinical & Diagnosis 2AFC & 9/12 (75.0\%) \\
 & Stroke Laterality & 2/9 (22.2\%) \\
 & LLPU 2AFC & 7/8 (87.5\%) \\
 & LLPU Laterality & 4/6 (66.7\%) \\
 & Amputation Level & 5/6 (83.3\%) \\
 & Balance & 9/20 (45.0\%) \\
Activity & Activity 2AFC & 24/31 (77.4\%) \\
 & Cross-Activity & 2/2 (100.0\%) \\
 & Walking Type & 11/14 (78.6\%) \\
Segmentation & Phase Duration & 12/15 (80.0\%) \\
 & Phase Timing & 15/20 (75.0\%) \\
 & Phase Count & 2/10 (20.0\%) \\
\bottomrule
\end{tabular}
\label{tbl-main-detailed}
\end{table}

Table~\ref{tbl-main-detailed} provides a detailed breakdown of performance by topic group within each category. Overall this shows that our baseline model performed well across many different tasks. For example, performance was near-perfect for database access tasks (98\%), with only the Data Shape topic showing any errors. The visualization tasks showed consistent performance generating plots aligned to requests (96\%), while perception tasks---which require interpreting the generated plots---showed more variable performance (66\%) reflecting the added difficulty of visual interpretation. Both task types used an LLM-as-a-Judge to determine if the response was consistent with the instructions, with the judge able to see the generated images.

\subsection{Spatiotemporal Performance}


\begin{table}
\centering
\caption[]{GaitTransformer ablation by spatiotemporal topic}

\begin{tabular}{p{\dimexpr 0.250\linewidth-2\tabcolsep}p{\dimexpr 0.250\linewidth-2\tabcolsep}p{\dimexpr 0.250\linewidth-2\tabcolsep}p{\dimexpr 0.250\linewidth-2\tabcolsep}}
\toprule
Topic & Baseline & +GaitTransformer & $\Delta$ \\
\hline
Event Counting & 0/8 (0.0\%) & 6/8 (75.0\%) & +75\% \\
Temporal & 0/12 (0.0\%) & 10/12 (83.3\%) & +83\% \\
Gait Events & 1/8 (12.5\%) & 4/8 (50.0\%) & +38\% \\
Velocity \& Length & 3/12 (25.0\%) & 9/12 (75.0\%) & +50\% \\
Asymmetry & 1/10 (10.0\%) & 3/10 (30.0\%) & +20\% \\
\bottomrule
\end{tabular}
\label{tbl-spatiotemporal-detailed}
\end{table}

We found that one domain in which BiomechAgent performed poorly was the detailed identification of gait event timing and the computation of spatiotemporal gait metrics. This is somewhat unsurprising, as performing this with only kinematic data would typically require designing and optimizing numerous heuristics to produce a reliable function. However, by giving BiomechAgent access to our GaitTransformer as an additional tool to analyze kinematic data and produce validated gait event timing, performance on these tasks improved markedly.

Table~\ref{tbl-spatiotemporal-detailed} presents a detailed breakdown of various topics, including gait analysis, compared with ground-truth data from our instrumented walkway, which the agent could not access. Measuring step length asymmetry was the topic that performed most poorly. Sources of performance limitations in these topics included differences between the GaitTransformer timing and the instrumented walkway, or the agent not using the exact same definitions of terms, such as step length, as the instrumented walkway. This requires privileged access to the mat's coordinate frame, although our prompts encourage analysis of heel positions relative to the direction of pelvis movement to approximate this.

\subsection{No Vision Performance}

We anticipated that using the vision capabilities of Gemini 2.5 Flash Lite to inspect plots would improve the performance and our custom instructions encouraged the model to perform this as a typical part of analyzing data when appropriate. This was motivated by how impressed we were at the ability for it to use this to visually estimate things like range of motion and gait event timing, as well as papers showing that visualizing graphs is a better representation for VLMs than just directly feeding time series data in as text \citep{liu_picture_2025}. To test this, we conducted a lesion experiment in which we disabled the transfer of generated plots to Gemini's visual inputs and removed instructions that would otherwise prompt it to attend to plots. As shown in Figure~\ref{fig-radar-comparison}, performance on most tasks was comparable or even better without visualization. This included tasks that we anticipated would benefit from graphing, such as clinical reasoning.

To understand when vision genuinely helps, we compared performance on two visual task categories: \textit{Perception tasks} that require interpreting generated plots (e.g., ``describe the movement pattern you observe'', ``estimate cadence from hip flexion peaks'') and \textit{Visualization tasks} that only require creating a correct plot (e.g., ``plot the knee angle during walking''). As shown in Figure~\ref{fig-radar-comparison}, vision capabilities helped with perception tasks (66\% with vision vs 44\% without), while visualization performance remained high regardless of vision (96\% vs 94\%). However, the model without vision was still able to write code that could solve many perception tasks, such as computing statistics or detecting peak locations in waveforms.

\subsection{Token Usage and Coding Error Rate}

\begin{table}
\centering
\caption[]{Efficiency metrics by model}

\begin{tabular}{p{\dimexpr 0.250\linewidth-2\tabcolsep}p{\dimexpr 0.250\linewidth-2\tabcolsep}p{\dimexpr 0.250\linewidth-2\tabcolsep}p{\dimexpr 0.250\linewidth-2\tabcolsep}}
\toprule
Model & Avg Time (s) & Avg Tokens & Avg Errors \\
\hline
Baseline & 12.4 & 61,365 & 0.58 \\
No Instructions & 10.0 & 39,575 & 0.27 \\
No Vision & 12.0 & 55,216 & 0.52 \\
MedGemma & 496.3 & 52,006 & 1.06 \\
\bottomrule
\end{tabular}
\label{tbl-efficiency}
\end{table}

In addition to accuracy, other important metrics for capturing our agent's behavior include the time to complete each task, the number of tokens used, and the number of coding errors produced. Table~\ref{tbl-efficiency} shows that all of our models used a consistent number of tokens, other than the model with no instructions, which used fewer. This model also completed the tasks more quickly, likely because our instructions encouraged additional steps, such as articulating hypotheses and reviewing results. The local model (MedGemma 27B) required two orders of magnitude more time per task and produced more than twice as many errors.


\subsection{Better Results with Improved Prompts}

The quality of the instructions, documentation of the tools, and the questions played an important role in the performance of BiomechAgent. With zero documentation of the tools, the agent would have no concept of how to use them, and performance would be at zero. The ablation results above show that our custom instructions contributed positively to performance. Our goal in developing our dataset was to create questions that were sufficiently detailed and clear to provide a reasonable test for agentic analysis of biomechanical data, without being overly detailed to the point of providing the answer in the question. For tests on which the agent failed, we often found that a more detailed or prescriptive question, which a human user might try, could yield a correct result. At one level, this reflects a realistic type of iterative chat an end user might have with the agent. At another extreme, it is moving all of the `intelligence' from the LLM to the questioner.

\subsubsection{Example 1: Hurdle Detection}

We asked the agent to identify which overground walking trial for a participant contained a hurdle step-over event. The original prompt provided no guidance on which kinematic features to examine: ``Through kinematic analysis of ALL the overground walking trials, please determine which trial [contained a hurdle step].'' The agent searched for explicit annotations labeled ``hurdle'' or ``obstacle,'' found none, and did not attempt kinematic analysis. After adding ``stepping over a hurdle requires greater hip flexion than normal walking,'' the agent computed maximum hip flexion across all trials, identified a clear outlier, and correctly identified the hurdle trial.

\subsubsection{Example 2: Stroke Detection}

We asked the agent to identify which of the two participants had a stroke based on gait kinematics. The original prompt asked to ``analyze joint angle patterns'' without specifying methodology. The agent focused on walking speed and provided the wrong answer. With methodological guidance specifying (1) compare only overground walking trials, (2) calculate left-right ROM asymmetry, and (3) use a \textgreater 15\% asymmetry threshold, the agent computed hip asymmetry of 17.15\% and knee asymmetry of 52.83\%, and correctly identified the participant with a stroke.

\section{Discussion}

In general, we found BiomechAgent to be a very powerful and convenient tool for accessing and visualizing data. BiomechAgent has become a regular first-line tool for us accessing and analyzing data, allowing us to `dogfood' this tool \citep{harrison_dogfood_2006} and convinced us of its robustness and utility. At the same time, the variety of uses presented a challenge for systematic benchmarking. Consequently, our quantitative benchmark covers only a small fraction of the task space that BiomechAgent might be able to accomplish. In our supplemental files, we included an HTML file that allows browsing transcripts of BiomechAgent interactions, which demonstrates the range of reasoning capabilities that emerge even on this finite dataset.

The consistent and robust performance of BiomechAgent in both data retrieval and plotting addresses a common request from clinicians using our markerless motion capture system. The ability to ask in natural language, `please pull up the knee kinematics during a single leg squat for participant ABC and compare the first session to the last session,' lowers the barrier to working with biomechanical data. This complements the benefit of markerless motion capture by lowering the barrier to data collection. While collecting and processing data to obtain kinematics is fairly straightforward, further analysis to generate clinical insights becomes the rate-limiting factor. This motivated the development of BiomechAgent, and the current capabilities already address many use cases. However, many important uses are not well addressed by BiomechAgent. For example, developing dashboards that longitudinally trend a standard set of metrics - such as gait parameters - in an intuitive and standardized manner is an everyday use case that is a poor fit for BiomechAgent. While it may be able to write an analysis, apply it to several trials, and extract results, it will perform a different analysis the next time the question is asked.

Our results on the spatiotemporal tasks are particularly noteworthy as they show an explicit limitation of frontier models on solving certain biomechanical tasks without access to despoke tools -- in this case, our GaitTransformer model that is itself another deep learning model trained to identify gait events \citep{cotton_transforming_2022, cimorelli_validation_2024}. This also raises the question of when additional bespoke analysis code should be handcrafted and validated, and then provided to BiomechAgent. In preliminary experiments, we found that, with guidance, BiomechAgent can produce reusable functions for tasks such as activity segmentation. We anticipate it will not be a big leap to extend BiomechAgent to develop its own tools and save them for future use, particularly when there is additional data available to validate this process. This is a natural extension of the growing use of coding agents in other aspects of programming. We have also demonstrated that BiomechAgent can easily and naturally use other tools such as BiomechGPT \citep{yang_biomechgpt_2025}, and when this tool in the agent prompt, it will naturally use this to solve tasks like activity and diagnosis classification, similarly, when provided access to KinTwin \citep{cotton_kintwin_2025}, it will use this to compute torques and ground reaction forces and to detect gait events when the GaitTransformer is unavailable.

We also envision that BiomechAgent will be just one agent in a zoo of agentic systems for analyzing rehabilitation data to deliver better care. For example, we have piloted agents that can extract structured rehabilitation information from therapy notes. These can be coordinated by supervising agents, who then run many such analyses across different individuals and time points, and identify and test novel hypotheses about rehabilitation, such as within our Causal Framework for Precision Rehabilitation \citep{cotton_causal_2024}. Such a system has been termed an AI Co-Scientist \citep{lu_ai_2024, gottweis_towards_2025}, and we hope BiomechAgent will be a powerful tool to accelerate quantitative movement analysis, providing actionable information for treatment optimization.

In addition to performing less accurately overall, running our benchmark with MedGemma 3 is relatively slow, even on an A100. Furthermore, a high-capacity (80GB) card is necessary to accommodate the long conversation lengths required for some analyses. Even then, some evaluations failed due to finite memory and exceeding the 128k maximum context length.  If performing large-scale analysis rather than running a chat interface, substantial speedups can be achieved by running multiple agents in parallel using tools such as vLLM \citep{kwon_efficient_2023}, as LLM inference scales efficiently with parallelization. This could be useful in situations where an organization is unable to use cloud-based language models, particularly if it is also analyzing relationships between other protected health information and biomechanics. With more transcripts of successful BiomechAgent usage, a smaller model could likely be fine-tuned for this range of biomechanical analysis tasks to achieve equivalent or better performance. An even more promising opportunity of using local models is training a combined BiomechGPT+Agent that can `perceive' biomechanics with System 1 reasoning, while then also writing analytic code to verify and refine those perceptions to provide robustness, grounding, and interpretability. Our benchmarks provide a large quantity of examples with known ground truth that can be used to evaluate the quality of the reasoning and open the opportunity for reinforcement learning from verifiable feedback \citep{deepseek-ai_deepseek-r1_2025, li_know_2025}. This can also be combined with more recent methods from Agent0 \citep{xia_agent0_2025}, which use another agent to develop a curriculum of more difficult challenges to advance the next-generation BiomechGPT+Agent.

The substantially better performance with Gemini 2.5 Flash Lite over MedGemma shows the importance of the `intelligence' in the LLM for solving these tasks. It serves to indicate that we are in a transitional period in which these models move from frequently failing on many tasks to reliably solving them. We attempted to evaluate the Gemini 3.0 Flash Preview but found the preview model's API unstable. It is also essential to consider security, regulatory requirements, and privacy when using a cloud-based LLM with clinical data. In our case, several features make this secure and compliant. Firstly, the database access tools provided to BiomechAgent grant access only to anonymous subject IDs, trial information, and kinematics (joint angles and locations). It is not able to access any of the videos used in the markerless motion capture analysis to produce these kinematics. Such kinematics are not considered identifiable information, and the NIH encourages and even mandates public sharing of them for federally funded research. Second, the kinematics do not actually leave the on-premises server. Instead, the LLM writes code to analyze the data and execute it on the server. However, in some cases, it may `print' kinematic data as text and is often encouraged to visualize plots of these data (i.e., Figure~\ref{fig-chat-interface}); such data is sent to the cloud. Thirdly, we have a Business Association Agreement (BAA) with our cloud LLM provider, which further governs the location of this data. However, we strongly encourage all readers to review their own applicable regulations and to use local models only when there is any uncertainty.

BiomechAgent is not without its limitations. It uses a fair number of tokens to answer questions, and our entire dataset of 310 questions consumed approximately 15-20M tokens.  Unsurprisingly, it still has limits; for vaguely phrased queries or questions that are not readily answered by visualizations, it is less likely to produce accurate responses. More specific questions, along with instructions encouraging it to generate hypotheses and reason them through by analyzing the data, helped improve performance. We anticipate there remain many opportunities to improve the performance and robustness to question phrasing through further refinement of the instructions, tool documentation, including the use of dedicated tools for this (e.g., DSPy \citep{singhvi_dspy_2023, khattab_dspy_2023}), and of course, the opportunity to fine-tune custom models for this. However, we defer this to further systematic end-user testing to delineate better real-world uses and failure modes. Finally, our analysis of spatiotemporal gait parameters shows that it is sometimes better to provide additional well-validated tools for a more detailed analysis

\section{Conclusion}

BiomechAgent demonstrates that code-generating AI agents can provide clinicians with accessible, natural language interfaces for biomechanical analysis, achieving robust accuracy across multiple tasks, including data retrieval, visualization, and clinical reasoning. By combining frontier language models with domain-specific instructions and specialized analysis tools, such systems can bridge the gap between the increasing availability of motion capture data and the need for clinician-friendly analysis.

\section{Acknowledgments}

We used Claude Code (Anthropic) during the development of BiomechAgent for software implementation, to assist in developing the evaluation dataset and topics from transcripts of chat usage, to help format the results into the manuscript, to ensure the code and the methods description were accurately aligned, and to support writing this manuscript. The actual core agent code was written only with the use of Github Copilot. Grammarly was used for editing.

This work was supported by R01HD114776 to RJC from National Institute of Child Health and Human Development and the National Center for Medical Rehabilitation Research.




{\small
\printbibliography
}

\end{document}